\documentclass[10pt,twocolumn,letterpaper]{article}

\usepackage{iccv}
\usepackage{times}
\usepackage{epsfig}
\usepackage{graphicx}
\usepackage{amsmath}
\usepackage{amssymb}
\usepackage{subcaption}

\usepackage{booktabs}
\usepackage[table]{xcolor}
\definecolor{lightgray}{gray}{0.9}
\definecolor{lightblue}{rgb}{0.93,0.95,1.0}
\definecolor{darkgreen}{rgb}{0.0,0.6,0.0}
\definecolor{mypink1}{rgb}{0.858, 0.188, 0.478}

\newcommand{\minisection}[1]{\vspace{2mm}\noindent{\textbf{#1}.}}

\usepackage[breaklinks=true,bookmarks=false]{hyperref}

\iccvfinalcopy 

\def\iccvPaperID{2619} 

\ificcvfinal\fi

\begin{document}

\makeatletter
\let\@oldmaketitle\@maketitle
\renewcommand{\@maketitle}{\@oldmaketitle
    \centering{\vspace{-5mm}\includegraphics[width=.95\linewidth,clip,trim = 0mm 0mm 0mm 0mm]{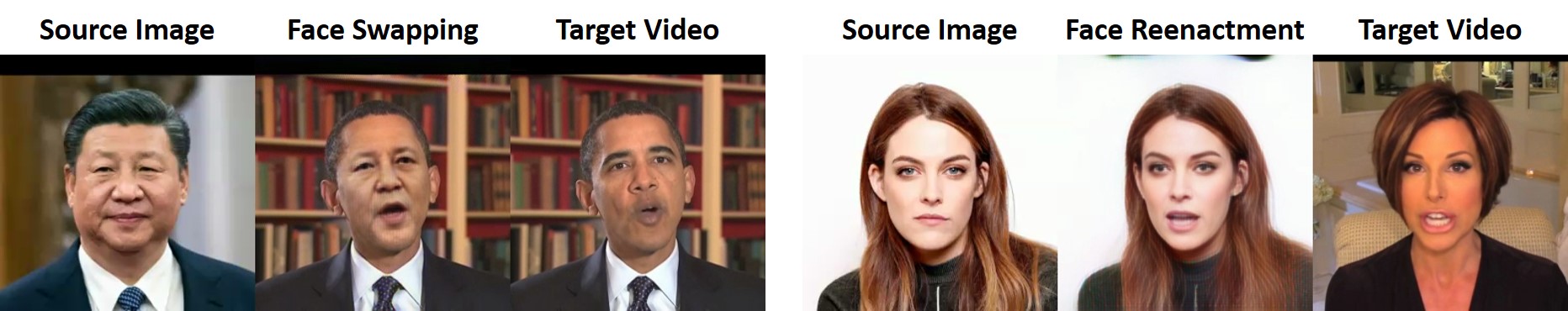}}
    \captionof{figure}{{\em Face swapping and reenactment.} Left: Source face swapped onto target. Right: Target video used to control the expressions of the face appearing in the source image. In both cases, our results appears in the middle. For more information please visit our website:~\url{https://nirkin.com/fsgan}.}\vspace{2mm}
    \label{fig:teaser}
    }
\makeatother

\title{FSGAN: Subject Agnostic Face Swapping and Reenactment}

\author{Yuval Nirkin\\
Bar-Ilan University, Israel\\
{\tt\small yuval.nirkin@gmail.com}
\and
Yosi Keller\\
Bar-Ilan University, Israel\\
{\tt\small yosi.keller@gmail.com}
\and
Tal Hassner\\
The Open University of Israel, Israel\\
{\tt\small talhassner@gmail.com}
}

\maketitle
\ificcvfinal\thispagestyle{empty}\fi

\begin{abstract}
We present Face Swapping GAN (FSGAN) for face swapping and reenactment. Unlike previous work, FSGAN is subject agnostic and can be applied to pairs of faces without requiring training on those faces. To this end, we describe a number of technical contributions. We derive a novel recurrent neural network (RNN)--based approach for face reenactment which adjusts for both pose and expression variations and can be applied to a single image or a video sequence. For video sequences, we introduce continuous interpolation of the face views based on reenactment, Delaunay Triangulation, and barycentric coordinates. Occluded face regions are handled by a face completion network. Finally, we use a face blending network for seamless blending of the two faces while preserving target skin color and lighting conditions. This network uses a novel Poisson blending loss which combines Poisson optimization with perceptual loss. We compare our approach to existing state-of-the-art systems and show our results to be both qualitatively and quantitatively superior.
\end{abstract}


\section{Introduction}

{\em Face swapping} is the task of transferring a face from source to target image, so that it seamlessly replaces a face appearing in the target and produces a realistic result (Fig.~\ref{fig:teaser} left). {\em Face reenactment} (aka {\em face transfer} or {\em puppeteering}) uses the facial movements and expression deformations of a control face in one video to guide the motions and deformations of a face appearing in a video or image (Fig.~\ref{fig:teaser} right). Both tasks are attracting significant research attention due to their applications in entertainment~\cite{alexander2009creating,kemelmacher2016transfiguring,wolf2010eye}, privacy~\cite{blanz2004exchanging,lin2012face,mosaddegh2014photorealistic}, and training data generation. 

Previous work proposed either methods for swapping or for reenactment but rarely both. Earlier methods relied on underlying 3D face representations~\cite{tran2017extreme} to transfer or control facial appearances. Face shapes were either estimated from the input image~\cite{thies2016face2face,suwajanakorn2017synthesizing,nirkin2018face} or were fixed~\cite{nirkin2018face}. The 3D shape was then aligned with the input images~\cite{chang17fpn} and used as a proxy when transferring intensities (swapping) or controlling facial expression and viewpoints (reenactment). 

Recently, deep network--based methods were proposed for face manipulation tasks. Generative adversarial networks (GANs)~\cite{goodfellow2014generative}, for example, were shown to successfully generate realistic images of fake faces. Conditional GANs (cGANs)~\cite{mirza2014conditional,isola2017image,wang2018pix2pixHD} were used to transform an image depicting real data from one domain to another and inspired multiple face reenactment schemes~\cite{pumarola2018ganimation,wayne2018reenactgan,sanchez2018triple}. Finally, the DeepFakes project~\cite{DeepFakes} leveraged cGANs for face swapping in videos, making swapping widely accessible to non-experts and receiving significant public attention. Those methods are capable of generating realistic face images by replacing the classic graphics pipeline. They all, however, still implicitly use 3D face representations.

Some methods relied on latent feature space domain separation~\cite{tian2018cr,natsume2018rsgan,natsume18fsnet}. These methods decompose the identity component of the face from the remaining traits, and encode identity as the manifestation of latent feature vectors, resulting in significant information loss and limiting the quality of the synthesized images. Subject specific methods~\cite{suwajanakorn2017synthesizing,DeepFakes,wayne2018reenactgan,kim2018deep} must be trained for each subject or pair of subjects and so require expensive subject specific data---typically thousands of face images---to achieve reasonable results, limiting their potential usage. Finally, a major concern shared by previous face synthesis schemes, particularly the 3D based methods, is that they all require special care when handling partially occluded faces.

We propose a deep learning--based approach to face swapping and reenactment in images and videos. Unlike previous work, our approach is {\em subject agnostic}: it can be applied to faces of different subjects without requiring subject specific training. Our Face Swapping GAN (FSGAN) is end-to-end trainable and produces photo realistic, temporally coherent results. We make the following contributions:
\begin{itemize}
    \item {\bf Subject agnostic swapping and reenactment.} To the best of our knowledge, our method is the first to simultaneously manipulate pose, expression, and identity without requiring person-specific or pair-specific training, while producing high quality and temporally coherent results.
    \item {\bf Multiple view interpolation.} We offer a novel scheme for interpolating between multiple views of the same face in a continuous manner based on reenactment, Delaunay Triangulation and barycentric coordinates.
    \item {\bf New loss functions.} We propose two new losses: A stepwise consistency loss, for training face reenactment progressively in small steps, and a Poisson blending loss, to train the face blending network to seamlessly integrate the source face into its new context.
\end{itemize}

We test our method extensively, reporting qualitative and quantitative ablation results and comparisons with state of the art. The quality of our results surpasses existing work even without training on subject specific images. 


\section{Related work}\label{sec:related_work}
Methods for manipulating the appearances of face images, particularly for face swapping and reenactment, have a long history, going back nearly two decades. These methods were originally proposed due to privacy concerns~\cite{blanz2004exchanging,lin2012face,mosaddegh2014photorealistic} though they are increasingly used for recreation~\cite{kemelmacher2016transfiguring} or entertainment (e.g.,~\cite{alexander2009creating,wolf2010eye}). 

\minisection{3D based methods}
The earliest swapping methods required manual involvement~\cite{blanz2004exchanging}. An automatic method was proposed a few years later~\cite{bitouk2008face}. More recently, Face2Face transferred expressions from source to target face~\cite{thies2016face2face}. Transfer is performed by fitting a 3D morphable face model (3DMM)~\cite{blanz2002face,blanz2003face,chang2019deep} to both faces and then applying the expression components of one face onto the other with care given to interior mouth regions. The reenactement method of Suwajanakorn et al.~\cite{suwajanakorn2017synthesizing} synthesized the mouth part of the face using a reconstructed 3D model of (former president) Obama, guided by face landmarks, and using a similar strategy for filling the face interior as in Face2Face. The expression of frontal faces was manipulated by Averbuch-Elor et al.~\cite{averbuch2017bringing} by transferring the mouth interior from source to target image using 2D wraps and face landmarks. 

Finally, Nirkin et al.~\cite{nirkin2018face} proposed a face swapping method, showing that 3D face shape estimation is unnecessary for realistic face swaps. Instead, they used a fixed 3D face shape as the proxy~\cite{hassner2015effective,masi2019face}. Like us, they proposed a face segmentation method, though their work was not end-to-end trainable and required special attention to occlusions. We show our results to be superior than theirs.


\minisection{GAN-based methods} GANs~\cite{goodfellow2014generative} were shown to generate fake images with the same distribution as a target domain. Although successful in generating realistic appearances, training GANs can be unstable and restricts their application to low-resolution images. Subsequent methods, however, improved the stability of the training process~\cite{mao2017least,arjovsky2017wasserstein}. Karras et al.~\cite{karras2017progressive} train GANs using a progressive multiscale scheme, from a low to high image resolutions. CycleGAN~\cite{zhu2017unpaired} proposed a cycle consistency loss, allowing training of unsupervised generic transformations between different domains. A cGAN with $L_{1}$ loss was applied by Isola et al.~\cite{isola2017image} to derive the pix2pix method, and was shown to produce appealing synthesis results for applications such as transforming edges to faces.

\minisection{Facial manipulation using GANs} Pix2pixHD~\cite{wang2018pix2pixHD} used GANs for high resolution image-to-image translation by applying a multi-scale cGAN architecture and adding a perceptual loss~\cite{johnson2016perceptual}. GANimation~\cite{pumarola2018ganimation} proposed a dual generator cGAN conditioned on emotion action units, that generates an attention map. This map was used to interpolate between the reenacted and original images, to preserve the background. GANnotation~\cite{sanchez2018triple} proposed deep facial reenactment driven by face landmarks. It generates images progressively using a triple consistency loss: it first frontalizes an image using landmarks then processes the frontal face.

Kim et al.~\cite{kim2018deep} recently proposed a hybrid 3D/deep method. They render a reconstructed 3DMM of a specific subject using a classic graphic pipeline. The rendered image is then processed by a generator network, trained to map synthetic views of each subject to photo-realistic images.

Finally, feature disentanglement was proposed as a means for face manipulation. RSGAN~\cite{natsume2018rsgan} disentangles the latent representations of face and hair whereas FSNet~\cite{natsume18fsnet} proposed a latent space which separates identity and geometric components, such as facial pose and expression.


\section{Face swapping GAN}\label{sec:FSGAN}

\begin{figure*}[!htbp]
\centering
\includegraphics[width=1.0\textwidth]{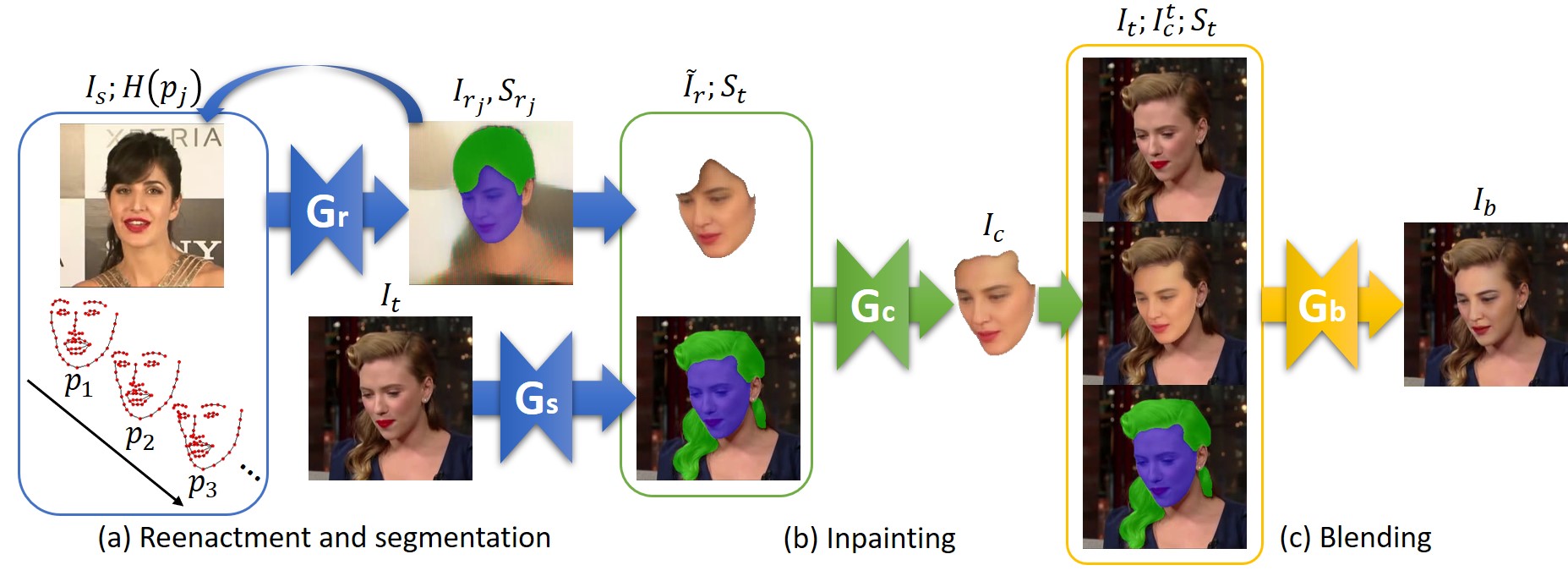}
\caption{Overview of the proposed FSGAN approach. (a) The recurrent reenactment generator $G_{r}$ and the segmentation generator $G_{s}$. $G_{r}$ estimates the reenacted face $F_{r}$ and its segmentation $S_r$, while $G_{s}$ estimates the face and hair segmentation mask $S_{t}$ of the target image $I_{t}$. (b) The inpainting generator $G_{c}$ inpaints the missing parts of $\tilde{F}_{r}$ based on $S_{t}$ to estimate the complete reenacted face $F_{c}$. (c) The blending generator $G_{b}$ blends $F_{c}$ and $F_{t}$, using the segmentation mask $S_{t}$.\vspace{-3mm}}\label{fig:system}
\end{figure*}

In this work we introduce the Face Swapping GAN (FSGAN), illustrated in Fig.~\ref{fig:system}. Let $I_{s}$ be the source and $I_{t}$ the target images of faces $F_{s}\in I_{s}$ and $F_{t}\in I_{t}$, respectively. We aim to create a new image based on $I_{t}$, where $F_{t}$ is replaced by $F_{s}$ while retaining the same pose and expression. 

FSGAN consists of three main components. The first, detailed in Sec.~\ref{subsec:Reenactment and Segmentation} (Fig.~\ref{fig:system}(a)), consists of a reenactment generator $G_r$ and a segmentation CNN $G_s$. $G_r$ is given a heatmaps encoding the facial landmarks of $F_{t}$, and generates the reenacted image ${I}_{r}$, such that $F_{r}$ depicts $F_{s}$ at the same pose and expression of $F_{t}$. It also computes
$S_{r}$: the segmentation mask of $F_{r}$. Component $G_s$
computes the face and hair segmentations of $F_{t}$.

The reenacted image, $I_r$, may contain missing face
parts, as illustrated in Fig.~\ref{fig:system} and Fig.~\ref{fig:system}(b). We therefore
apply the face inpainting network, $G_c$, detailed in Sec.~\ref{subsec:inpainting}
using the segmentation $S_{t}$, to estimate the missing pixels. The final part of the FSGAN, shown in Fig.~\ref{fig:system}(c) and Sec.~\ref{subsec:Blending}, is the blending of the completed face $F_{c}$ into the target image $I_{t}$ to derive the final face swapping result.

The architecture of our face segmentation network, $G_{s}$, is based on U-Net~\cite{ronneberger2015u}, with bilinear interpolation for upsampling. All our other generators---$G_{r}$, $G_{c}$, and $G_{b}$---are based on those used by pix2pixHD~\cite{wang2018pix2pixHD}, with coarse-to-fine generators and multi-scale discriminators. Unlike pix2pixHD, our global generator uses a U-Net architecture with bottleneck blocks~\cite{he2016deep} instead of simple convolutions and summation instead of concatenation. As with the segmentation network, we use bilinear interpolation for upsampling in both global generator and enhancers. The actual number of layers differs between generators. 

Following others~\cite{wayne2018reenactgan}, training subject agnostic face reenactment is non-trivial and might fail when applied to unseen face images related by large poses. To address this challenge, we propose to break large pose changes into
small manageable steps and interpolate between the closest available source images corresponding to a target's pose. These steps are explained in the following sections.

\subsection{Training losses}

\label{subsec:Training-Losses}

\minisection{Domain specific perceptual loss} 
To capture fine facial details we adopt the perceptual loss~\cite{johnson2016perceptual}, widely used in recent work for
face synthesis~\cite{sanchez2018triple}, outdoor scenes~\cite{wang2018pix2pixHD}, and super resolution~\cite{ledig2017photo}. Perceptual loss uses the feature maps of a pretrained VGG network, comparing high frequency details using a Euclidean distance.

We found it hard to fully capture details inherent to face images, using a network pretrained on a generic dataset such as ImageNet. Instead, our network is trained on the target
domain: We therefore train multiple VGG-19 networks~\cite{simonyan2014very} for face recognition and face attribute classification. Let $F_{i}\in
\mathbb{R}
^{C_{i}\times H_{i}\times W_{i}}$ be the feature map of the $i$-th layer of
our network, the perceptual loss is given by%
\begin{equation}
\mathcal{L}_{perc}(x,y)=\sum_{i=1}^{n}\frac{1}{C_{i}H_{i}W_{i}}\left\Vert
F_{i}(x)-F_{i}(y)\right\Vert _{1}.\label{eq:perceptualloss}
\end{equation}

\minisection{Reconstruction loss} 
While the perceptual loss of Eq.~\eqref{eq:perceptualloss}
captures fine details well, generators trained using only that loss, often produce images with inaccurate colors, corresponding to reconstruction of low frequency image content. We hence also applied a pixelwise $L_{1}$ loss to the generators:%
\begin{equation}
\mathcal{L}_{pixel}(x,y)=\Vert x-y\Vert _{1}.
\end{equation}%
The overall loss is then given by
\begin{equation}
\mathcal{L}_{rec}(x,y)=\lambda _{perc}\mathcal{L}_{perc}(x,y)+\lambda
_{pixel}\mathcal{L}_{pixel}(x,y).  \label{equ:total loss}
\end{equation}
The loss in Eq.~\eqref{equ:total loss} was used with all our generators.%

\minisection{Adversarial loss} 
To further improve the realism of our generated images we use an adversarial objective~\cite{wang2018pix2pixHD}. We utilized a multi-scale discriminator consisting of multiple discriminators, $D_{1},D_{2},...,D_{n}$, each one operating on a different image resolution. For a generator $G$ and a multi-scale discriminator $D$, our adversarial loss is defined by:
\begin{equation}
\mathcal{L}_{adv}(G,D)=\min_{G}\max_{D_{1},\dots D_{n}}\sum_{i=1}^{n}%
\mathcal{L}_{GAN}(G,D_{i}),
\end{equation}%
where $\mathcal{L}_{GAN}(G,D)$ is defined as:
\begin{align}
\mathcal{L}_{GAN}(G,D)=&\mathbb{E}_{(x,y)}[\log D(x,y)]\nonumber\\&+\mathbb{E}_{x}[\log
(1-D(x,G(x)))].
\end{align}

\subsection{Face reenactment and segmentation}
\label{subsec:Reenactment and Segmentation}
Given an image $I\in\mathbb{R}^{3\times H\times W}$ and a heatmap
representation $H(p)\in\mathbb{R}^{70\times H\times W}$ of facial landmarks,
$p\in\mathbb{R}^{70\times2}$, we define the face reenactment generator, $G_{r}$,
as the mapping $G_{r}:\left\{  \mathbb{R}^{3\times H\times W},\mathbb{R}%
^{70\times H\times W}\right\}  \rightarrow\mathbb{R}^{3\times H\times W}$. 

Let
$v_{s},v_{t}\in\mathbb{R}^{70\times3}$ and $e_{s},e_{t}\in\mathbb{R}^{3}$, be
the 3D landmarks and Euler angles corresponding to $F_{s}$ and $F_{t}$. We generate intermediate 2D landmark positions $p_{j}$ by interpolating between $e_{s}$ and $e_{t}$, and the centroids of $v_{s}$ and $v_{t}$, using intermediate points for which we project $v_{s}$ back to $I_{s}$. We define the reenactment
output recursively for each iteration $1\leq j\leq n$ as
\begin{equation}
I_{r_{j}},S_{r_{j}}=G_{r}(I_{r_{j-1}};H(p_{j})),
\end{equation}
$$I_{r_{0}}=I_{s}.$$

Similar to others~\cite{pumarola2018ganimation}, the last layer of the global generator and each of the enhancers in $G_{r}$ is split into two heads: the first produces the reenacted image and the second the segmentation mask. In contrast to binary masks used bu others~\cite{pumarola2018ganimation}, we consider the face and hair regions separately. The binary mask implicitly
learned by the reenactment network captures most of the head including the hair, which we segment separately. Moreover, the additional hair segmentation also improves the accuracy of the face segmentation. The face segmentation generator $G_{s}$ is defined as $G_{r}:\mathbb{R}^{3\times H\times
W}\rightarrow\mathbb{R}^{3\times H\times W}$, where given an RGB image it output a 3-channels segmentation mask encoding the background, face, and hair.

\minisection{Training}
Inspired by the triple consistency loss~\cite{sanchez2018triple}, we propose a stepwise consistency loss. Given an image pair $(I_s,I_t)$ of the same subject from a video sequence, let $I_{r_n}$ be the reenactment result after $n$ iterations, and $\widetilde{I_t},\widetilde{I}_{r_n}$ be the same images with their background removed using the segmentation masks $S_t$ and $S_{r_j}$, respectively. The stepwise consistency loss is defined as: $\mathcal{L}_{rec}(\widetilde{I}_{r_n},\widetilde{I}_t)$.
The final objective for the $G_r$:
\begin{align}
\mathcal{L}(G_r)=&\lambda_{stepwise}\mathcal{L}_{rec}(\widetilde{I}_{r_n},\widetilde{I}_t)+\lambda_{rec}\mathcal{L}_{rec}(\widetilde{I}_r,\widetilde{I}_t)\nonumber\\&+\lambda_{adv}\mathcal{L}_{adv}+\lambda_{seg}\mathcal{L}_{pixel}(S_r,S_t).
\end{align}


For the objective of $G_s$ we use the standard cross-entropy loss, $L_{ce}$, with additional guidance from $G_r$:
\begin{equation}
\mathcal{L}(G_s)=L_{ce}+\lambda_{reenactment}\mathcal{L}_{pixel}(S_t,S_r^t),
\end{equation}
where $S_r^t$ is the segmentation mask result of $G_r(I_t;H(p_t))$ and $p_t$ is the 2D landmarks corresponding to $I_t$.

We train both $G_r$ and $G_s$ together, in an interleaved fashion. We start with training $G_s$ for one epoch followed by the training of $G_r$ for an additional epoch, increasing $\lambda_{reenactment}$ as the training progresses. We have found that training $G_r$ and $G_s$ together helps filtering noise learned from coarse face and hair segmentation labels.

\subsection{Face view interpolation}
\label{subsec:FaceViewInterpolation}
\begin{figure*}[t]
\centering
\includegraphics[clip,trim=0mm 3mm 0mm 0mm, width=.85\textwidth]{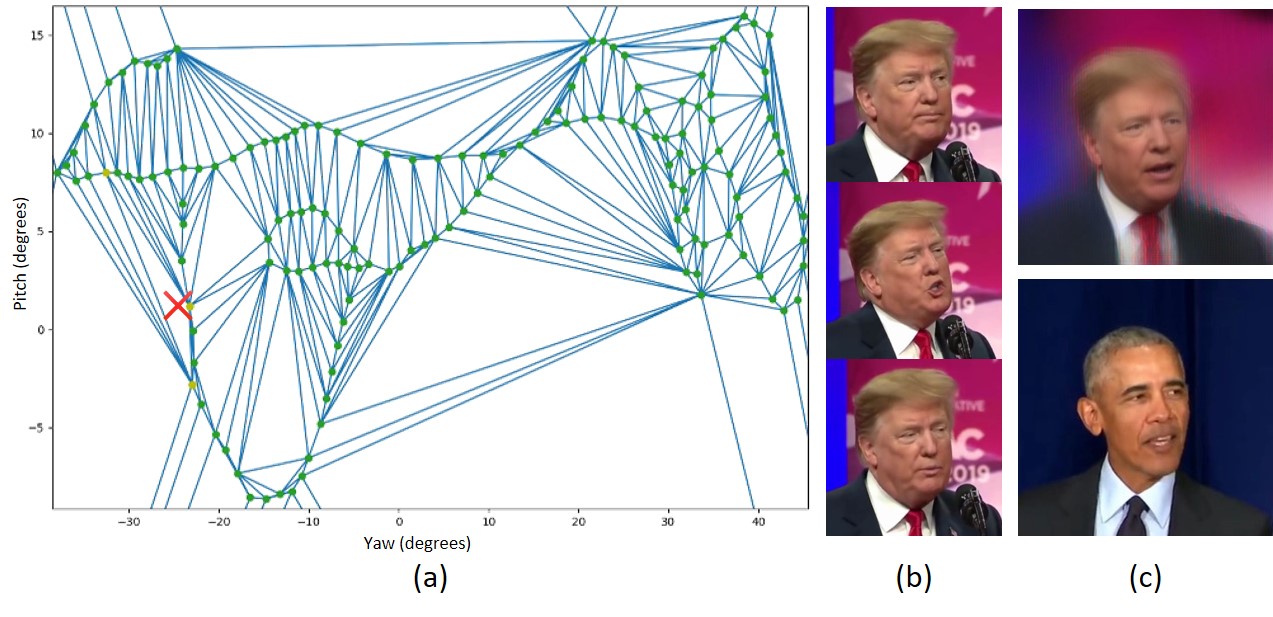}\vspace{-3mm}
\caption{
{\em Face view interpolation.} (a) Shows an example of an appearance map of the source subject (Donald Trump). The green dots represent different views of the source subject, the blue lines represent the Delaunay Triangulation of those views, and the red X marks the location of the current target's pose. (b) The interpolated views associated with the vertices of the selected triangle (represented by the yellow dots). (c) The reenactment result and the current target image. \vspace{-4mm} }\label{fig:interpolation}
\end{figure*}
Standard computer graphics pipelines project textured mesh polygons onto a plane for seamless rendering~\cite{hughes2014computer}. We propose a novel, alternative scheme for continuous interpolation between face views. This step is an essential phase of our method, as it allows using the entire source video sequence, without training our model on a particular video frame, making it subject agnostic.

Given a set of source subject images, $\left\{ \mathbf{I}%
_{s_{1}},\dots ,\mathbf{I}_{s_{n}}\right\}$, and Euler angles, $\left\{
\mathbf{e}_{1},\dots ,\mathbf{e}_{n}\right\}$, of the corresponding faces $\left\{ \mathbf{F}_{s_{1}},\dots ,\mathbf{F}_{s_{n}}\right\}$, we construct
the appearance map of the source subject, illustrated in Fig.~\ref{fig:interpolation}(a). This appearance map embeds head poses in a triangulated plane, allowing head poses to follow continuous paths.

We start by projecting the Euler angles $\left\{ \mathbf{e}_{1},\dots ,\mathbf{e}_{n}\right\} $ onto a plane by dropping the roll angle. Using a k-d tree data structure~\cite{hughes2014computer}, we remove points in the angular domain that are too close to each other, prioritizing the points for which the corresponding Euler angles have a roll angle closer to zero. We further remove motion blurred images. Using the remaining points, $\left\{ x_{1},\dots ,x_{m}\right\}$, and the four boundary points, $y_{i}\in \lbrack -75,75]\times \lbrack -75,75]$, we build a
mesh, $M$, in the angular domain by Delaunay Triangulation.

For a query Euler angle, $e_{t}$, of a face, $F_{t}$, and its corresponding projected point, $x_{t}$, we find the triangle $T\in M$ that contains $x_{t}$. Let $x_{i_{1}},x_{i_{2}},x_{i_{3}}$ be the vertices of $T$ and $I_{s_{i_{1}}},I_{s_{i_{2}}},I_{s_{i_{3}}}$ be the corresponding face views. We calculate the barycentric coordinates, $\lambda _{1},\lambda _{2},\lambda
_{3}$ of $x_{t}$, with respect to $x_{i_{1}},x_{i_{2}},x_{i_{3}}$. The
interpolation result $I_{r}$ is then
\begin{equation}
I_{r}=\sum_{k=1}^{3}\lambda _{k}G_{r}(I_{s_{i_{k}}};H(\mathbf{p}_{t})),
\end{equation}
where $\mathbf{p}_{t}$ are the 2D landmarks of $F_{t}$. If any vertices of the triangle are boundary points, we exclude them from the interpolation and normalize the weights, $\lambda _{i}$, to sum to one. 

A face view query is illustrated in Fig.~\ref{fig:interpolation}(b,c). To improve interpolation accuracy, we use a horizontal flip to fill in views when the appearance map is one-sided with respect to the yaw dimension, and generate artificial views using $G_{r}$ when the appearance map is too sparse.

\subsection{Face inpainting}
\label{subsec:inpainting}
Occluded regions in the source face $F_{s}$ cannot be rendered on the target face, $F_{t}$. Nirkin et al.~\cite{nirkin2018face} used the segmentations of $F_{s}$ and $F_{t}$ to remove occluded regions, rendering (swapping) only regions visible in both source and target faces. 
Large occlusions and different facial textures can cause noticeable artifacts in the resulting images. 

To mitigate such problems, we apply a face inpainting generator, $G_{c}$ (Fig.~\ref{fig:system}(b)). $G_{c}$ renders face image $F_{s}$ such that the resulting face rendering $\tilde{I}_{r}$ covers entire segmentation mask $S_{t}$ (of $F_{t}$), thereby resolving such occlusion.

Given the reenactment result, $I_r$, its corresponding segmentation, $S_r$, and the target image with its background removed, $\tilde{I}_t$, all drawn from the same identity, we first augment $S_r$ by simulating common face occlusions due to hair, by randomly removing ellipse-shaped parts, in various sizes and aspect ratios from the border of $S_r$. Let $\tilde{I}_r$ be $I_r$ with its background removed using the augmented version of $S_r$, and $I_c$ the completed result from applying $G_c$ on $\tilde{I}_r$. We define our inpainting generator loss as
\begin{equation}
\mathcal{L}(G_{c})=\lambda _{rec}\mathcal{L}_{rec}(I_c,\tilde{I}_t)+\lambda _{adv}\mathcal{L}_{adv},
\end{equation}%
where $\mathcal{L}_{rec}$ and $\mathcal{L}_{adv}$ are the reconstruction and adversarial losses of Sec.~\ref{subsec:Training-Losses}.

\subsection{Face blending}
\label{subsec:Blending}
The last step of the proposed face swapping scheme is blending of the completed face $F_c$ with its target face $F_{t}$ (Fig. \ref{fig:system}(c)). Any blending must account for, among others, different skin tones and lighting conditions. Inspired by previous uses of Poisson blending for inpainting~\cite{yeh2017semantic} and blending~\cite{wu2017gp}, we propose a novel Poisson blending loss.

Let $I_{t}$ be the target image, $I_{r}^{t}$ the image of the reenacted face transferred onto the target image, and $S_{t}$ the segmentation mask marking the transferred pixels. Following~\cite{perez2003poisson}, we define the Poisson blending optimization as
\begin{equation}
\label{equ:Poisson}
   \begin{alignedat}{2}
    P(I_{t};I_{r}^{t};S_{t}))= & \arg \min_{f}\Vert \nabla f-\nabla I_{r}^{t}\Vert _{2}^{2} \\
    & \text{s.t. }f(i,j)=I_{t}(i,j),\text{ }\forall \text{ }S_{t}(i,j)=0,
   \end{alignedat}
\end{equation}
where $\nabla \left( \cdot \right) $ is the gradient operator. We combine the Poisson optimization in Eq.~\eqref{equ:Poisson} with the perceptual loss. The Poisson blending loss is then $\mathcal{L}(G_{b})$
$$
\mathcal{L}(G_{b})=\lambda _{rec}\mathcal{L}_{rec}(G_{b}(I_{t};I_{r}^{t};S_{t}),P(I_{t};I_{r}^{t};S_{t}))+\lambda_{adv}\mathcal{L}%
_{adv}.
$$


\section{Datasets and training}
\label{sec:datasets_and_processing}
\subsection{Datasets and processing}
We use the video sequences of the IJB-C dataset~\cite{maze2018iarpa} to train our generator, $G_{r}$, for which we automatically extracted the frames depicting particular subjects. IJB-C contains $\sim$11k face videos, of
which we used 5,500 which were in high definition. Similar to the frame pruning approach of Sec.~\ref{subsec:FaceViewInterpolation}, we prune the face
views that are too close together as well as motion-blurred frames. 

We apply the segmentation CNN, $G_{s}$, to the frames, and prune the frames for which less than 15\% of the pixels in the face bounding box were classified as face pixels. We used dlib's face verification\footnote{Available: \url{http://dlib.net/}} to group frames according to the subject identity, and limit the number of frames per subject to 100, by choosing frames with the maximal variance in 2D landmarks. In each training iteration, we choose the frames $I_{s}$ and $I_{t}$ from two randomly chosen subjects.

We trained VGG-19 CNNs for the perceptual loss on the VGGFace2 dataset~\cite{cao2018vggface2} for face recognition and the CelebA~\cite{liu2018large} dataset for face attribute classification. The VGGFace2 dataset contains 3.3M images depicting 9,131 identities, whereas CelebA contains 202,599 images, annotated with 40 binary attributes.

We trained the segmentation CNN, $G_{s}$, on data used by others~\cite{nirkin2018face}, consisting of ${\sim}10k$ face images labeled with face segmentations. We also used the LFW Parts Labels set~\cite{kae2013augmenting} with ${\sim}3k$ images labeled for face and hair segmentations, removing the neck regions using facial landmarks. 

We used additional 1k images and corresponding hair segmentations from the Figaro dataset~\cite{svanera2016figaro}. Finally, FaceForensics++~\cite{roessler2019faceforensics++} provides 1000 videos, from which they generated 1000 synthetic videos on random pairs using DeepFakes~\cite{DeepFakes} and Face2Face~\cite{thies2016face2face}.

\subsection{Training details}
We train the proposed generators from scratch, where the weights were initialized randomly using a normal distribution. We use Adam optimization~\cite{kingma2014adam} ($\beta_{1}=0.5,\beta_{2}=0.999$) and a learning rate of $0.0002$. We reduce this rate by half every ten epochs. The following parameters were used for all the generators: $\lambda_{perc}=1,\lambda_{pixel}=0.1,\lambda_{adv}=0.001,\lambda_{seg}=0.1,\lambda_{rec}=1,\lambda_{stepwise}=1$, where $\lambda_{reenactment}$ is linearly increased from 0 to 1 during training. All of our networks were trained on eight NVIDIA Tesla V100 GPUs and an Intel Xeon CPU. Training of $G_{s}$ required six hours to converge, while the rest of the networks converged in two days. All our networks, except for $G_{s}$, were trained using a progressive multi scale approach, starting with a resolution of 128$\times$128 and ending at 256$\times$256. Inference rate is ${\sim}30$fps for reenactment and ${\sim}10$fps for swapping on one NVIDIA Tesla V100 GPU.


\section{Experimental results}
We performed extensive qualitative and quantitative experiments to verify the proposed scheme. We compare our method to two previous face swapping methods: DeepFakes~\cite{DeepFakes} and Nirkin et al.~\cite{nirkin2018face}, and the Face2Face reenactment scheme~\cite{thies2016face2face}. We conduct all our experiments on videos from FaceForensics++~\cite{roessler2019faceforensics++}, by running our method on the same pairs they used. We further report ablation studies showing the importance of each component in our pipeline.

\subsection{Qualitative face reenactment results}
\begin{figure*}[t]
\centering
\includegraphics[width=1.0\linewidth,trim=0mm 0mm 0mm 0mm ,clip]{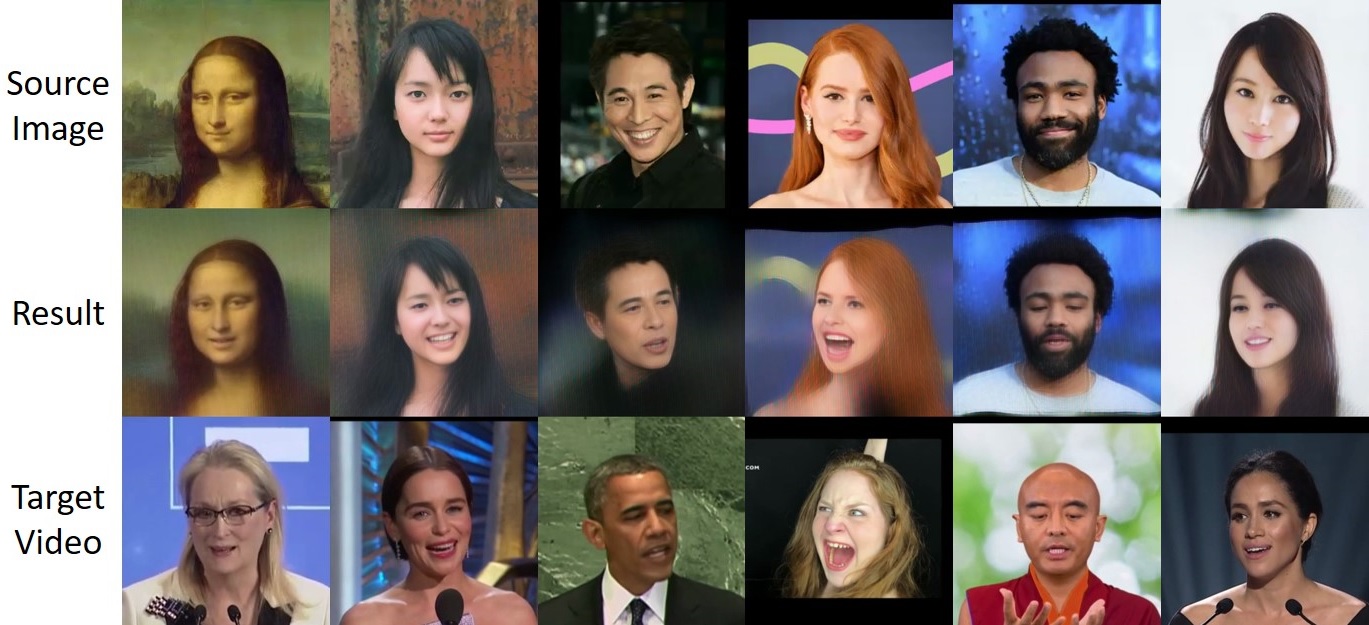}
\caption{ \emph{Qualitative face reenactment results.} Row 1: The source face for reenactment. Row 2: Our reenactment results (without background removal). Row 3: The target face from which to transfer the pose and expression.  \vspace{-4mm} }%
\label{fig:face_reenactment_qualitative}%
\end{figure*}

\begin{figure}[!htbp]
\centering
\includegraphics[width=1.0\linewidth]{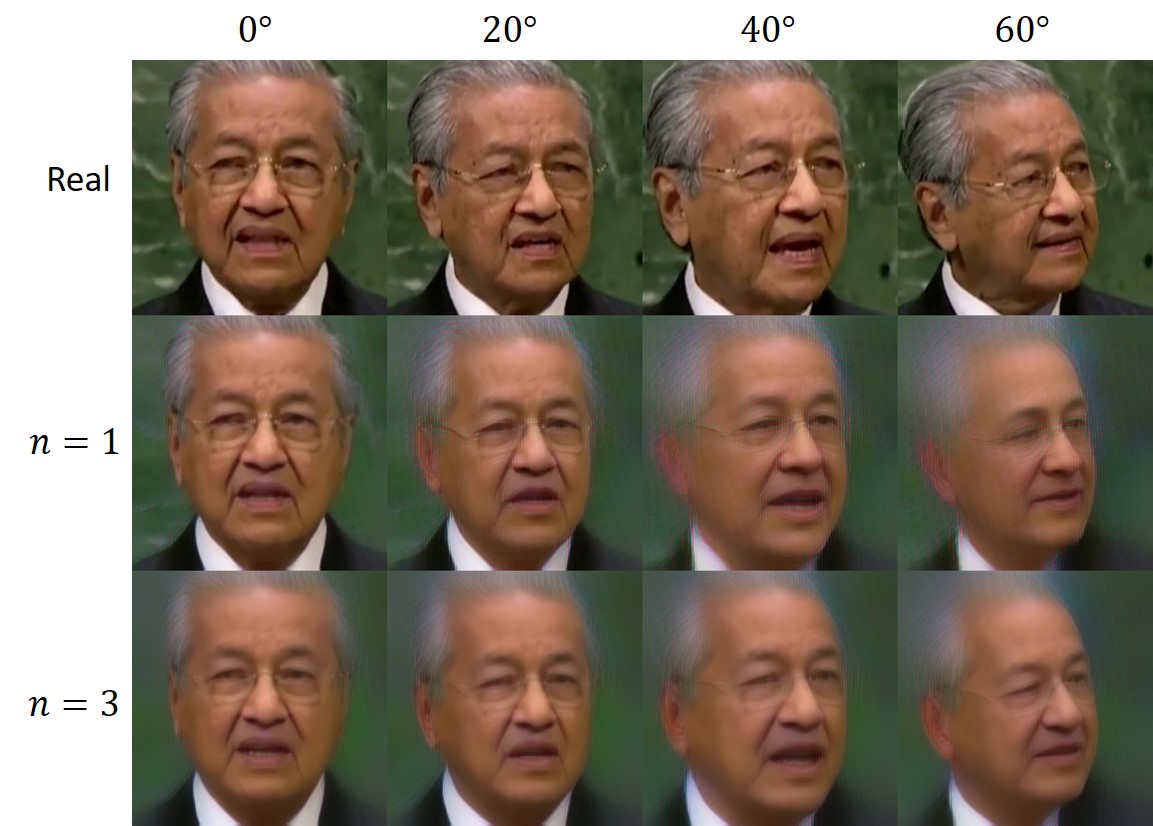}
\caption{ \emph{Reenactment limitations.} Top left image transformed onto each of the images in Row 1 (using the same subject for clarity). Row 2: Reenactment with one iteration. Row 3: Three iterations.}%
\label{fig:reenactment_limitations}\vspace{-5mm}%
\end{figure}

Fig.~\ref{fig:face_reenactment_qualitative} shows our raw face reenactment results, without background removal. We chose examples of varying ethnicity, pose, and expression. A specifically interesting example can be seen in the rightmost column, showing our method's ability to cope with extreme expressions. To show the importance of iterative reenactment, Fig~\ref{fig:reenactment_limitations} provides reenactments of the same subject for both small and large angle differences. As evident from the last column, for large angle differences, the identity and texture are better preserved using multiple iterations. 

\subsection{Qualitative face swapping results}
\begin{figure*}[tb]
\centering
\includegraphics[width=1.0\linewidth,trim=0mm 27mm 0mm 0mm ,clip]{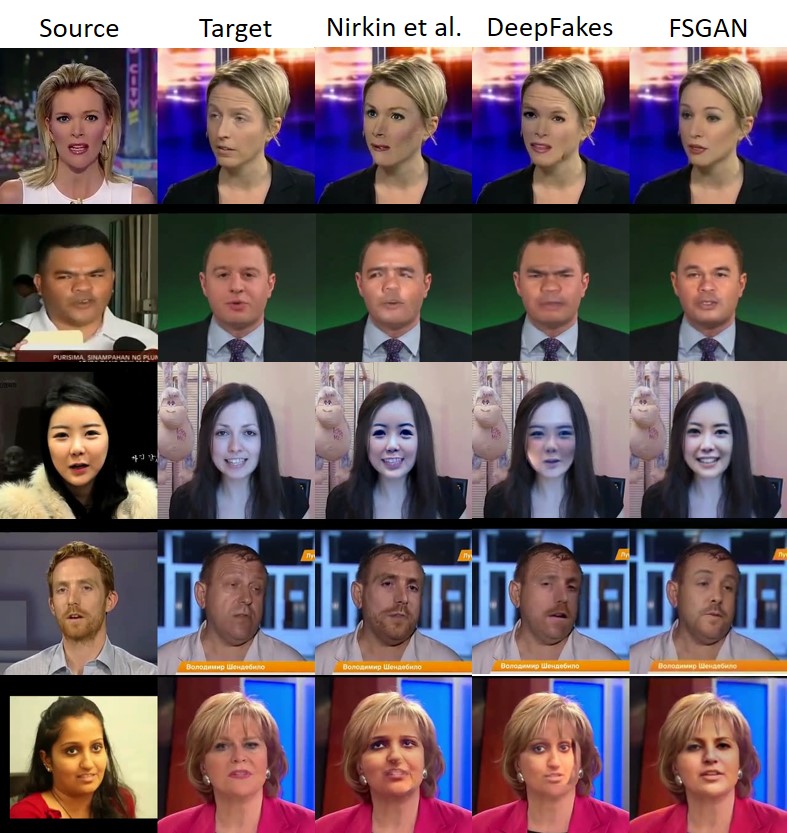}
\caption{ \emph{Qualitative face swapping results on~\cite{roessler2019faceforensics++}.} Results for source photo swapped onto target provided for Nirkin et al.~\cite{nirkin2018face}, DeepFakes~\cite{DeepFakes} and our method on images of faces of subjects it was not trained on.  }%
\label{fig:face_swap_qualitative}\vspace{-5mm}%
\end{figure*}

Fig.~\ref{fig:face_swap_qualitative} offers face swapping examples taken from FaceForensics++ videos, {\em without training our model on these videos}. We chose examples that represent different poses and expression, face shapes, and hair occlusions. Because Nirkin et al.~\cite{nirkin2018face} is an image-to-image face swapping method, to be fair in our comparison, for each frame in the target video we select the source frame with the most similar pose. To compare FSGAN in a video-to-video scenario, we use our face view interpolation described in Sec.~\ref{subsec:FaceViewInterpolation}.


\subsection{Comparison to Face2Face}
\begin{figure}[t]
\centering
\includegraphics[width=1.0\linewidth]{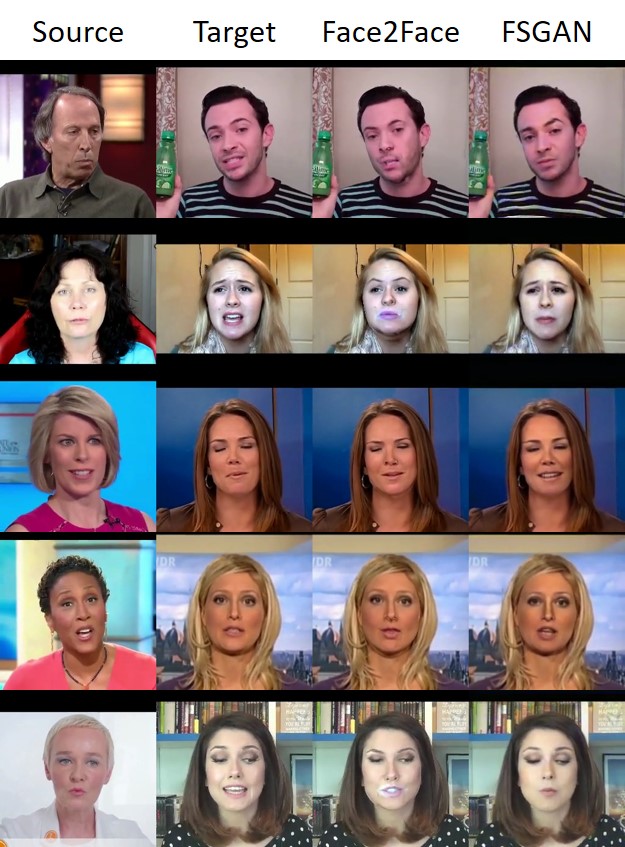}
\caption{ \emph{Comparison to Face2Face~\cite{thies2016face2face} on FaceForensics++~\cite{roessler2019faceforensics++}.} As demonstrated, our method exhibits far less artifacts than Face2Face. \vspace{-4mm} }%
\label{fig:face2face}%
\end{figure}

We compare our method to Face2Face~\cite{thies2016face2face} on the expression only reenactment problem. Given a pair of faces $F_s \in I_s$ and $F_t \in I_t$ the goal is to transfer the expression from $I_s$ to $I_t$. To this end, we modify the corresponding 2D landmarks of $F_t$ by swapping in the mouth points of the 2D landmarks of $F_s$, similarly to how we generate the intermediate landmarks in Sec.~\ref{subsec:Reenactment and Segmentation}. The reenactment result is then given by $G_r(I_t;H(\hat{p}_t))$, where $\hat{p}_t$ are the modified landmarks. The examples are shown in Fig.~\ref{fig:face2face}.

\begin{figure*}[!htbp]
\centering
\includegraphics[width=0.95\linewidth,trim=0mm 29mm 0mm 0mm ,clip]{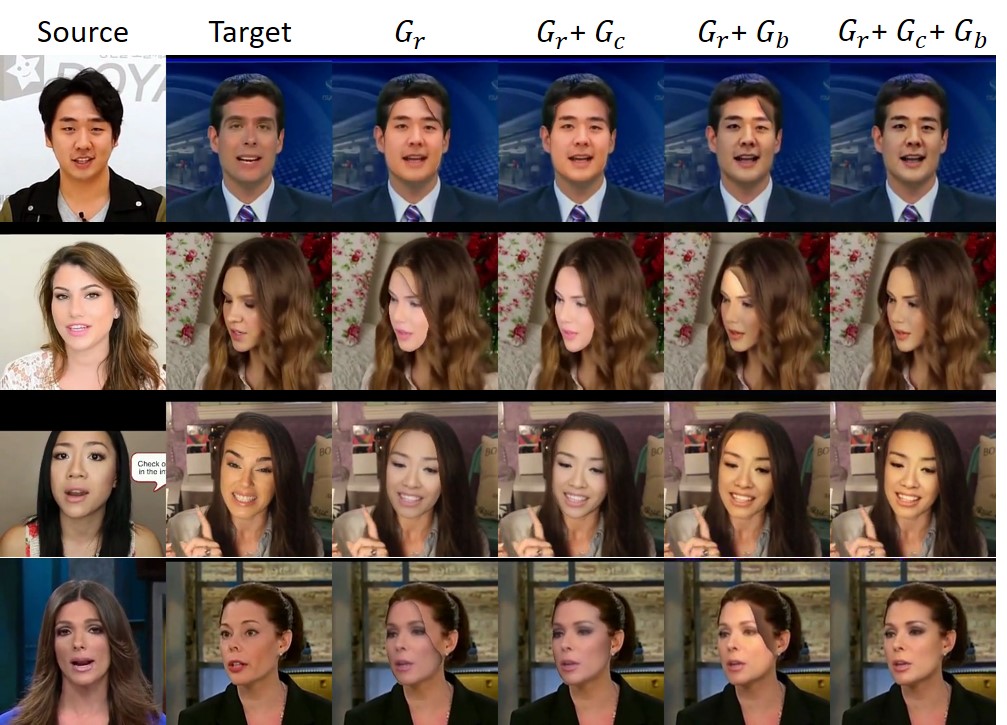}\vspace{-2mm}
\caption{\emph{Ablation study}. From columns 3 and 5, without the completion network, $G_c$, the transferred face does not cover the entire target face, leaving obvious artifacts. Columns 3 and 4 show that without the blending network, $G_b$, the skin color and lighting conditions of the transferred face are inconsistent with its new context.}%
\vspace{-4mm}
\label{fig:ablation_qualitative}%
\end{figure*}

\subsection{Quantitative results}
\label{subsec:Quantitative results}
We report quantitative results, conforming to how we defined the face swapping problem: we validate how well methods preserve the source subject identity, while retaining the same pose and expression of the target subject. To this end, we first compare the face swapping result, $F_b$, of each frame to its nearest neighbor in pose from the subject face views. We use the dlib~\cite{dlib09} face verification method to compare identities and the structural similarity index method (SSIM) to compare their quality. To measure pose accuracy, we calculate the Euclidean distance between the Euler angles of $F_b$ to the original target image, $I_t$. Similarly, the accuracy of the expression is measured as the Euclidean distance between the 2D landmarks. Pose error is measured in degrees and the expression error is measured in pixels. We computes the mean and variance of those measurements on the first 100 frames of the first 500 videos in FaceForensics++, averaging them across the videos. As baselines, we use Nirkin et al.~\cite{nirkin2018face} and DeepFakes~\cite{DeepFakes}. 

Evident from the first two columns of Table~\ref{tab:quant}, our approach preserves identity and image quality similarly to previous methods. The two rightmost metrics in Table~\ref{tab:quant} show that our method retains pose and expression much better than its baselines. Note that the human eye is very sensitive to artifacts on faces. This should be reflected in the quality score but those artifacts usually capture only a small part of the image and so the SSIM score does not reflect them well.

\begin{table}[t!]
\centering{
 \resizebox{0.98\linewidth}{!}{
 \begin{tabular}{lcccc} 
 \toprule
 Method & verification $\downarrow$ & SSIM $\uparrow$ & euler $\downarrow$ & landmarks $\downarrow$ \\ [0.5ex] 
 \hline
 Nirkin et al.~\cite{nirkin2018face} & 0.39 $\pm$ 0.00       & 0.49 $\pm$ 0.00       & 3.15 $\pm$ 0.04       & 26.5 $\pm$ 17.7 \\
 DeepFakes ~\cite{DeepFakes} & 0.38 $\pm$ 0.00 & 0.50 $\pm$ 0.00 & 4.05 $\pm$ 0.04 & 34.1 $\pm$ 16.6 \\ 
 FSGAN                       & 0.38 $\pm$ 0.00 & {\bf 0.51 $\pm$ 0.00} & {\bf 2.49 $\pm$ 0.04} & {\bf 22.2 $\pm$ 17.7} \\
 \bottomrule
 \end{tabular}
}
}
\caption{{\em Quantitative swapping results.} On FaceForensics++ videos~\cite{roessler2019faceforensics++}.}\label{tab:quant}\vspace{-4mm}
\end{table}

\subsection{Ablation study}

We performed ablation tests with four configurations of our method: $G_r$ only, $G_r+G_c$, $G_r+G_b$, and our full pipeline. The segmentation network, $G_s$, is used in all configurations. Qualitative results are provided in Fig.~\ref{fig:ablation_qualitative}.

Quantitative ablation results are reported in Table~\ref{tab:ablation_quant}. Verification scores show that source identities are preserved across all pipeline networks. From Euler and landmarks scores we see that target poses and expressions are best retained with the full pipeline. Error differences are not extreme, suggesting that the inpainting and blending generators, $G_{c}$ and $G_{b}$, respectively, preserve pose and expression similarly well. There is a slight drop in the SSIM, due to the additional networks and processing added to the pipeline.

\begin{table}[tbh]
\centering{
\resizebox{0.98\linewidth}{!}{
\begin{tabular}{lcccc}
\toprule
Method & verification $\downarrow$ & SSIM $\uparrow$ & euler $\downarrow$ & landmarks $\downarrow$ \\ [0.5ex] 
\hline
FSGAN $(G_r)$          & 0.38 $\pm$ 0.00 & 0.54 $\pm$ 0.00 & 3.16 $\pm$ 0.03 & 22.6 $\pm$ 16.5 \\
FSGAN $(G_r+G_c)$      & 0.38 $\pm$ 0.00 & 0.54 $\pm$ 0.00 & 3.21 $\pm$ 0.08 & 24.5 $\pm$ 17.2 \\
FSGAN $(G_r+G_b)$      & 0.38 $\pm$ 0.00 & 0.52 $\pm$ 0.00 & 2.75 $\pm$ 0.05 & 23.6 $\pm$ 17.9 \\
FSGAN $(G_r+G_c+G_b)$  & 0.38 $\pm$ 0.00 & 0.51 $\pm$ 0.00 & {\bf 2.49 $\pm$ 0.04} & {\bf 22.2 $\pm$ 17.7} \\
\bottomrule
\end{tabular}
}
}
\caption{{\em Quantitative ablation results.} On
FaceForensics++ videos~\cite{roessler2019faceforensics++}.}%
\label{tab:ablation_quant}\vspace{-5mm}%
\end{table}


\section{Conclusion}
\noindent{\bf Limitations.} Fig.~\ref{fig:reenactment_limitations} shows our reenactment results for different facial yaw angles. Evidently, the larger the angular differences, the more identity and texture quality degrade. Moreover, too many iterations of the face reenactment generator blur the texture. Unlike 3DMM based methods, e.g., Face2Face~\cite{thies2016face2face}, which warp textures directly from the image, our method is limited to the resolution of the training data. Another limitation arises from using a sparse landmark tracking method that does not fully capture the complexity of facial expressions. 

\minisection{Discussion} Our method eliminates laborious, subject-specific, data collection and model training, making face swapping and reenactment accessible to non-experts. We feel strongly that it is of {\em paramount importance} to publish such technologies, in order to drive the development of technical counter-measures for detecting such forgeries, as well as compel law makers to set clear policies for addressing their implications. Suppressing the publication of such methods would not stop their development, but rather make them available to select few and potentially blindside policy makers if it is misused.


{\small
\bibliographystyle{ieee_fullname}
\bibliography{egbib}
}


\onecolumn

\pagebreak

   \newpage
   \null
   \vskip .375in
   \begin{center}
      {\Large \bf Supplementary Material \par}
   \end{center}
\appendix

\section{Additional qualitative results}
\begin{figure}[ptb]
\centering
\includegraphics[width=\linewidth]{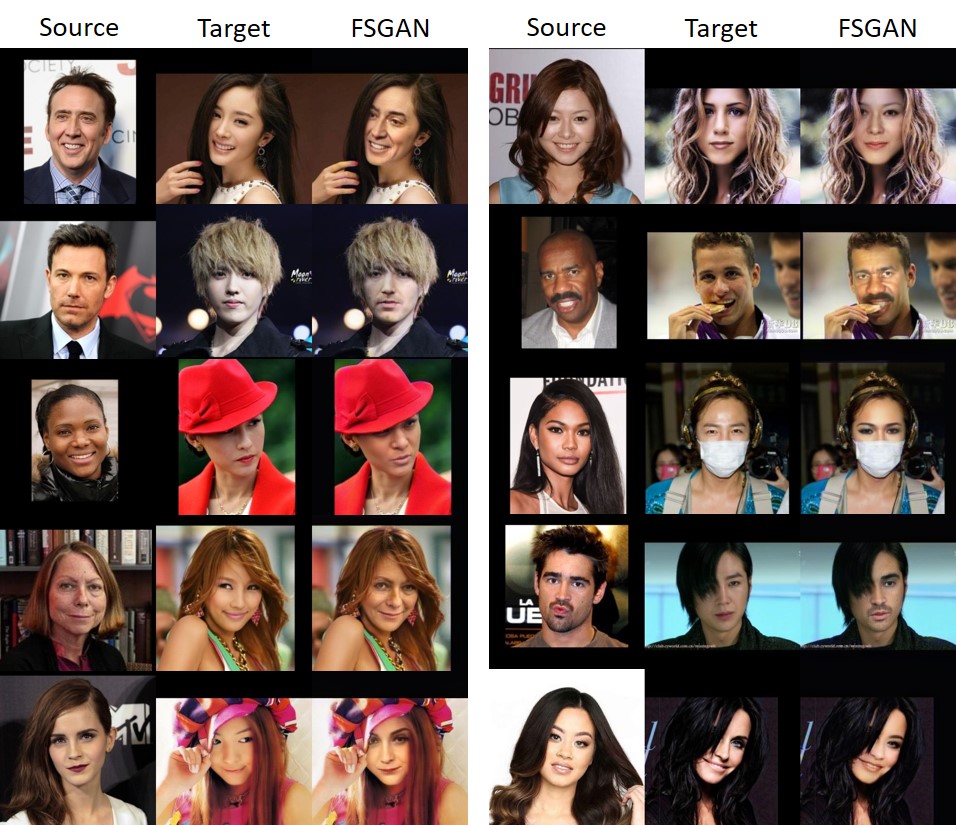}\caption{
\emph{Additional qualitative face swapping results on on the Caltech Occluded Faces in the Wild (COFW) dataset~\cite{burgos2013robust}.}}%
\label{fig:additional_quant}%
\end{figure}

\begin{figure}[ptb]
\centering
\includegraphics[width=\linewidth]{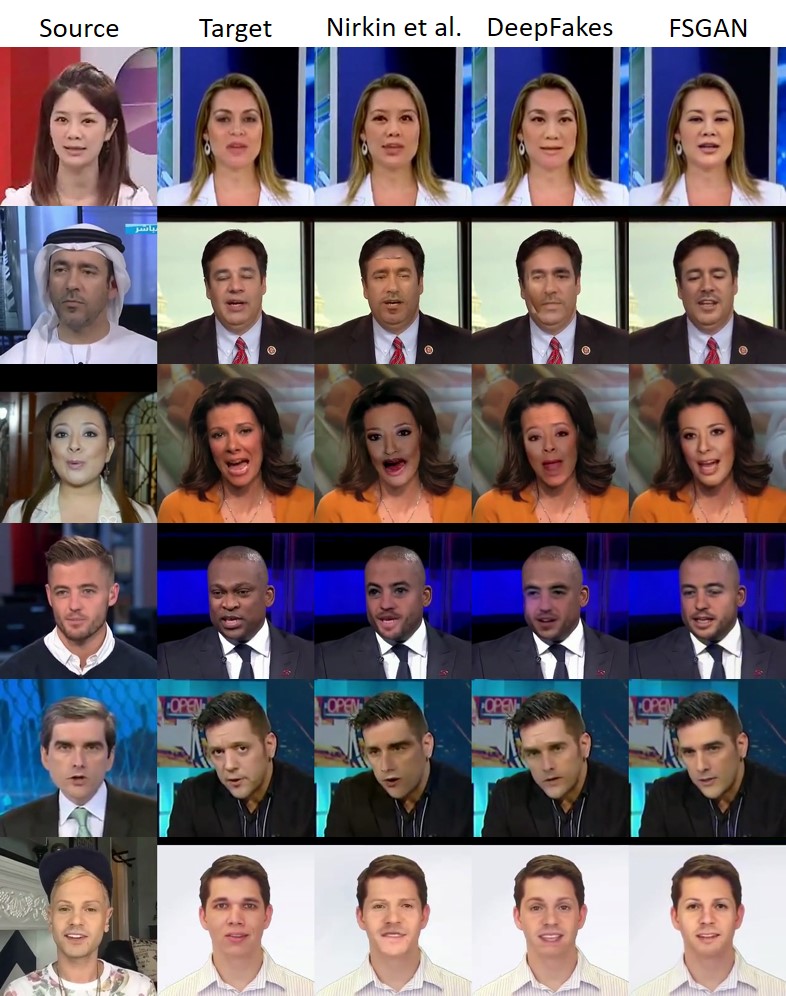}\caption{
\emph{Additional qualitative face swapping comparison to Nirkin et al.~\cite{nirkin2018face} and DeepFakes~\cite{DeepFakes} on FaceForensics++~\cite{roessler2019faceforensics++}.}}%
\label{fig:additional_comparison}%
\end{figure}

\begin{figure}[ptb]
\centering
\includegraphics[width=0.75\linewidth]{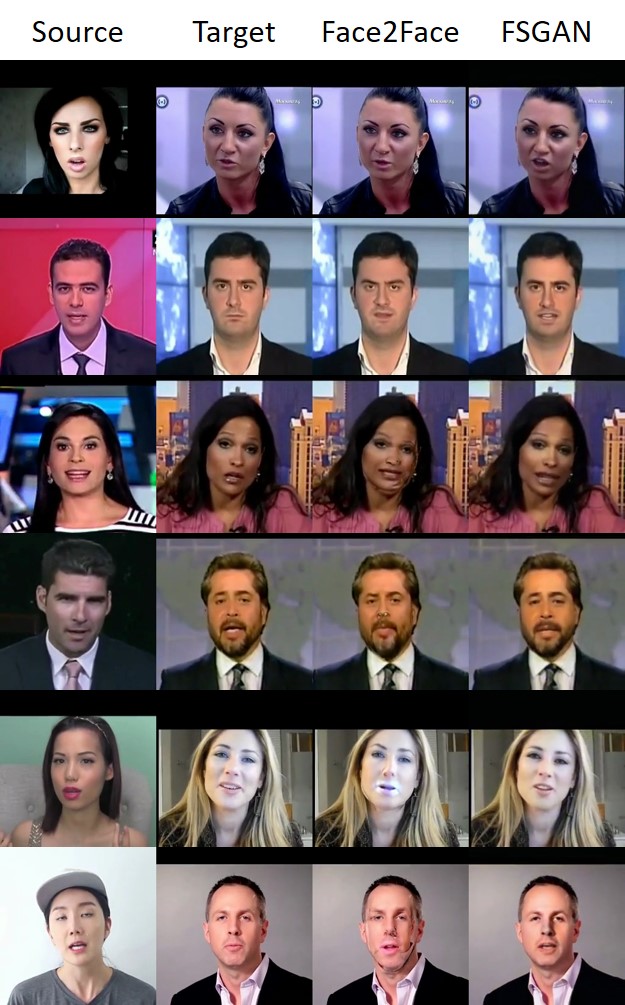}
\caption{\emph{Additional qualitative face reenactment comparison to Face2Face~\cite{thies2016face2face} on FaceForensics++~\cite{roessler2019faceforensics++}.}}%
\label{fig:additional_face2face}%
\end{figure}

We offer additional quantitative face swapping results in Fig.~\ref{fig:additional_quant}. We have specifically chosen examples of challenging pairs, with partial occlusions, different ethnicities and skin colors, demonstrating the competence of our method on a large variety of subjects. 
In Fig.~\ref{fig:additional_comparison}, we show additional quantitative comparison to Nirkin et al.~\cite{nirkin2018face} and DeepFakes~\cite{DeepFakes}, and in Fig.~\ref{fig:additional_face2face} we show another comparison to Face2Face~\cite{thies2016face2face}.
Please also see the attached video for more results.

\section{The architecture of the generator CNNs}
\label{sec:architecture}
\begin{figure}[ptb]
\centering
\includegraphics[width=\linewidth]{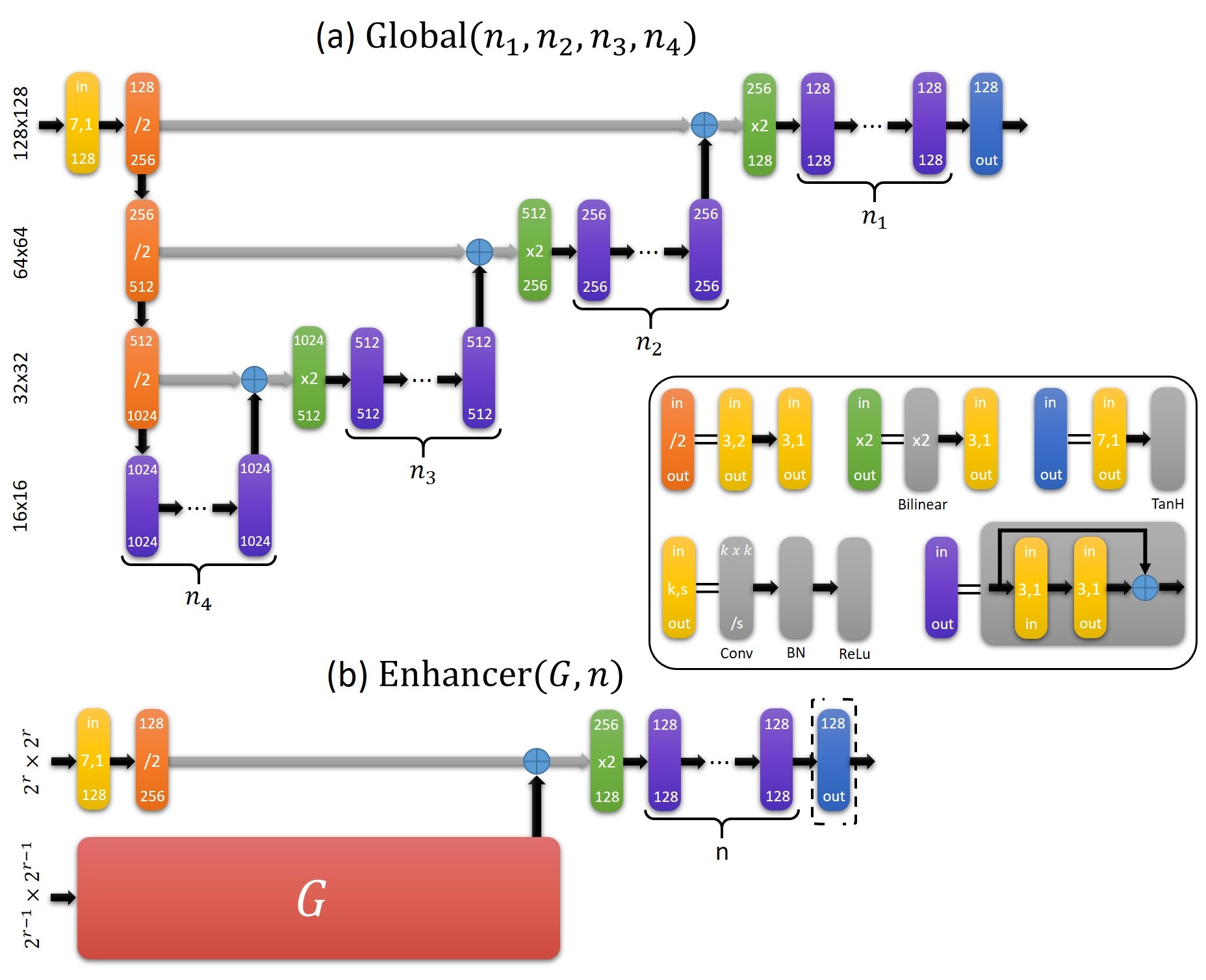}\caption{
\emph{Generator architectures.} (a) The global generator is based on a residual variant of the U-Net~\cite{ronneberger2015u} CNN, using a number of bottleneck layers per resolution. We replace the simple convolutions with bottleneck blocks (in purple), the concatenation with summation (plus sign), and the deconvolutions with bilinear upsamplnig following by a convolution. (b) The enhancer utilizes a submodule and a number of bottleneck layers. The last output block (in blue) is only used in the enhancer of the finest resolution.}%
\label{fig:architecture}%
\end{figure}

The architecture of the generators, $G_{r}$, $G_{c}$, and $G_{b}$, is based on the pix2pixHD approach~\cite{wang2018pix2pixHD}, and the layout of the global generator and enhancer is depicted in Fig.~\ref{fig:architecture}. The global generator is defined by the number of bottleneck blocks (shown in purple) used in each resolution scale. In our experiments we used only three resolutions. The enhancer is defined by its submodule, that is, either the global generator or another enhancer, and its number of bottleneck layers. The generators are thus given
by%
\[
G_{r}=G_{c}=\text{Enhancer}(\text{Global}(2,2,3),2),
\]
and
\[
G_{b}=\text{Enhancer}(\text{Global}(1,1,1),1).
\]
The face segmentation network $G_{s}$ is based on the U-Net~approach~\cite{ronneberger2015u}, for which we replaced the deconvolution layers with bilinear interpolation upsampling layers.

\end{document}